\documentclass[conference]{IEEEtran}
\IEEEoverridecommandlockouts
\usepackage{cite}
\usepackage{amsmath,amssymb,amsfonts}
\usepackage{algorithm}
\usepackage{algorithmic}
\usepackage{array}
\usepackage{booktabs}
\usepackage[nolist]{acronym}
\usepackage{graphicx}
\usepackage{textcomp}
\usepackage{xcolor}
\usepackage[T1]{fontenc}
\usepackage{caption}

\newcommand{\rVec}{\ensuremath{\boldsymbol{r}}}

\newcommand{\pc}{\ensuremath{\boldsymbol{p}_{c}}}
\newcommand{\pcd}{\ensuremath{\hat{\boldsymbol{p}}_{c}}}
\newcommand{\pca}{\ensuremath{\boldsymbol{p}_{c}^{a}}}

\newcommand{\tdi}{\ensuremath{\hat{\boldsymbol{t}}_{i}}}
\newcommand{\ti}{\ensuremath{\boldsymbol{t}_{i}}}
\newcommand{\tdj}{\ensuremath{\hat{\boldsymbol{t}}_{j}}}
\newcommand{\tj}{\ensuremath{\boldsymbol{t}_{j}}}

\newacro{poi}[POI]{Point of Interest}
\newacroplural{poi}[POIs]{Points of Interest}
\newacro{p2p}[P2P]{point-to-point}
\newacro{p2s}[P2S]{point-to-surface}
\newacro{pcg}[PCG]{preconditioned conjugate gradients}
\newacro{tsdf}[TSDF]{Truncated Signed Distance Field}

\def\BibTeX{{\rm B\kern-.05em{\sc i\kern-.025em b}\kern-.08em
    T\kern-.1667em\lower.7ex\hbox{E}\kern-.125emX}}
\begin{document}

\setlength{\abovedisplayskip}{4pt}
\setlength{\belowdisplayskip}{4pt}
\setlength{\belowcaptionskip}{-15pt}

\title{Localization and Tracking of User-Defined Points on Deformable Objects for Robotic Manipulation}

\author{\IEEEauthorblockN{Sven Dittus,
Benjamin Alt, Andreas Hermann, Darko Katic and Rainer Jäkel}
\IEEEauthorblockA{ArtiMinds Robotics, Karlsruhe, Germany\\
\{sven.dittus | benjamin.alt | andreas.hermann | darko.katic | rainer.jaekel\}@artiminds.com}\\
\IEEEauthorblockN{Jürgen Fleischer}
\IEEEauthorblockA{Institute of Production Science, Karlsruhe Institute of Technology, Germany \\
juergen.fleischer@kit.edu}
\thanks{This work was partially supported by the German Federal Ministry of Education and Research (BMBF) under the grant no. 16SV8406.}}

\maketitle

\global\csname @topnum\endcsname 0  
\global\csname @botnum\endcsname 0  

\begin{abstract}
This paper introduces an efficient procedure to localize user-defined points on the surface of deformable objects and track their positions in 3D space over time. To cope with a deformable object's infinite number of DOF, we propose a discretized deformation field, which is estimated during runtime using a multi-step non-linear solver pipeline. The resulting high-dimensional energy minimization problem describes the deviation between an offline-defined reference model and a pre-processed camera image. An additional regularization term allows for assumptions about the object’s hidden areas and increases the solver’s numerical stability. Our approach is capable of solving the localization problem online in a data-parallel manner, making it ideally suitable for the perception of non-rigid objects in industrial manufacturing processes.
\end{abstract}


\section{Introduction}
Many manufacturing processes rely on image processing to enable industrial robots to manipulate objects. Whereas many sophisticated camera systems meet the need for localizing and tracking user-defined \acp{poi} on rigid objects, there is still no sufficiently accurate solution for coping with this problem for deformable objects yet. Moreover, existing approaches reconstruct the deformable object‘s model online, requiring \acp{poi} to be defined at runtime and thus being unsuitable for fully automated processes. This paper proposes a solution for defining POIs on an offline model and then localizing and tracking these points on a deformable object in an online detection pipeline.

\begin{figure}
    \includegraphics[width=\linewidth]{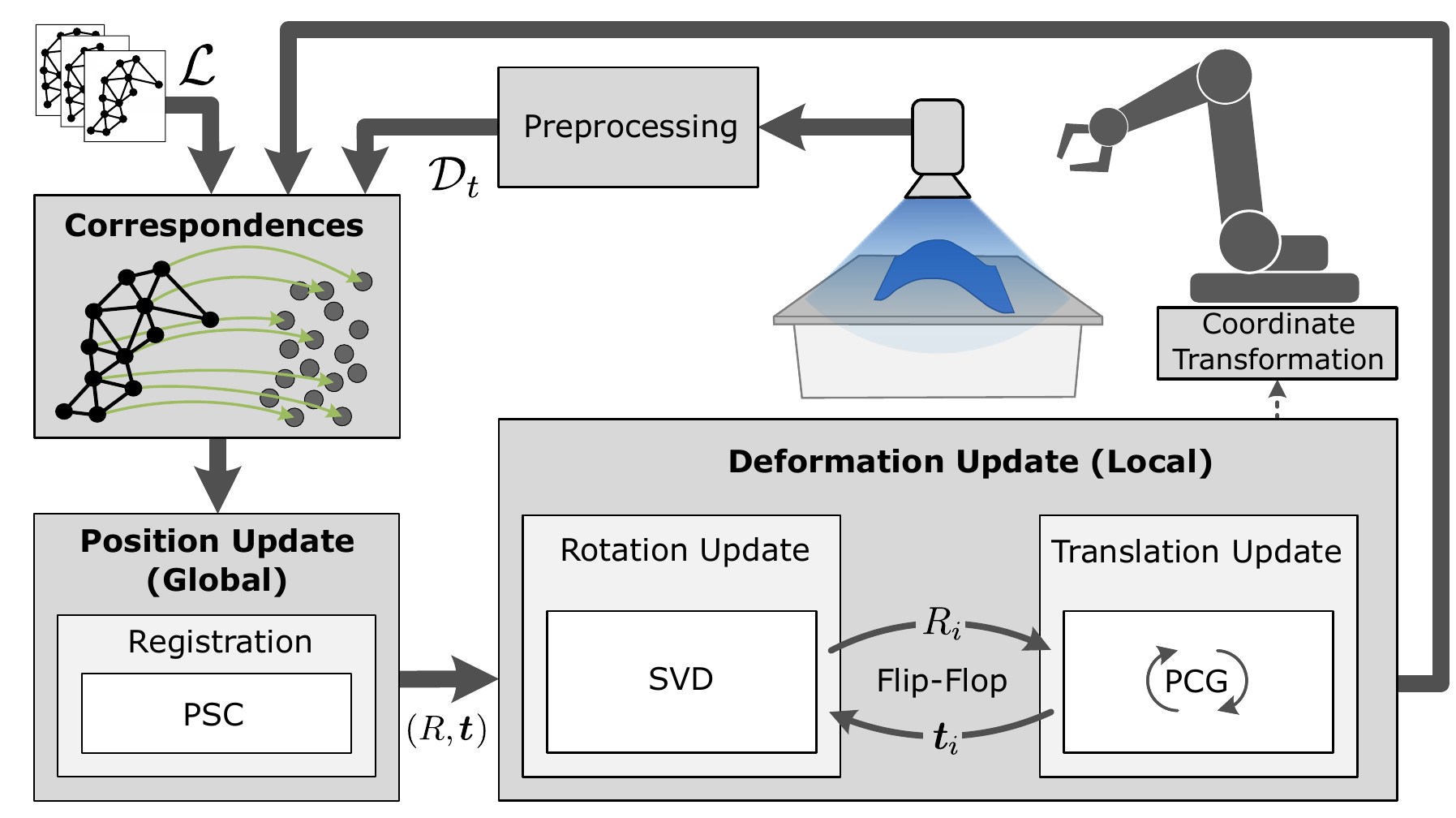}
    \caption{Overview of the localization and tracking pipeline}
\end{figure}

\section{Related Work}
Existing approaches for localizing and tracking \acp{poi} are only applicable to specific object categories (e.g. linear \cite{Tang.2018} or planar \cite{Tang.2013,Schulman.2013}), assume speficic deformation models (e.g. articulated models \cite{Schmidt.2015} or skeletons \cite{Gall.2009}) or particular materials (e.g. textiles \cite{Li.2014,Li.2018}) or are restricted to detecting specific features (e.g. points on corners or edges \cite{Yamazaki.2014,Ramisa.2012}). To reach the precision required for sophisticated manipulation tasks, prior work requires external markers \cite{Finnegan.2006,Trumble.2017} or elaborate physics models \cite{Bay.2006,Tian.2010,Lang.2011,Schulman.2013,Leizea.2014}. We combine several SotA-solutions for rigid object state estimation (such as SHOT descriptors), physical modelling (such as deformation fields) and computer graphics (such as projection) into a processing pipeline which permits the tracking and localization of user-defined points on arbitrary deformable objects using only a single depth camera without markers or prior physics modeling.

\section{Method}
Our algorithm consists of three distinct phases: (1)  \textbf{Demonstration} of the reference model and the \acp{poi}; (2) an \textbf{iterative localization and tracking process} consisting of \textbf{observing} a new point cloud, identification of \textbf{correspondences} between the observation and the deformed reference model of the previous timestep and \textbf{estimation of the deformation}; and (3) a \textbf{coordinate transformation} of the localized \acp{poi} into the robot end-effector coordinate system for subsequent manipulation.
\subsection{Surface and deformation model}
\label{sec:surface_and_deformation_model}
\begin{figure}
    \centering
    \includegraphics[height=5.5em]{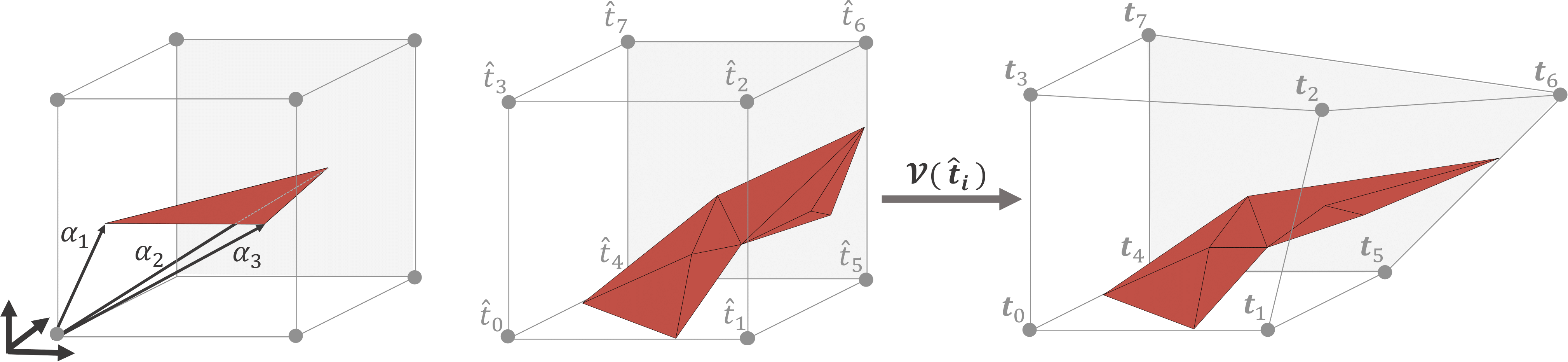}
    \caption{Trilinear weights (l.); static and deformed grid cells (r.)}
    \label{fig:deformation_model}
\end{figure}

To efficiently perform computations, we model object surfaces as triangle meshes, while deformations are modelled via a \textit{deformation grid} \cite{Innmann.2016}. Unlike \cite{Innmann.2016}, we use a highly detailed mesh as a surface representation which is independent of the deformation model's resolution. This allows to increase computation performance while maintaining a highly detailed surface. 
Our deformation model consists of two data structures, both containing $\lvert G \rvert$ grid points. Whereas the equally spaced static grid $G$ describes the undeformed reference model, the deformation field $\mathcal{V}$ represents the object's deformed state at the current timestep $t$. Each gridpoint $i$ is defined in $G$ by a position vector $\tdi$, allowing to express the position $\hat{\boldsymbol{p}}$ of an undeformed vertex within $G$  as $\hat{\boldsymbol{p}} = \sum^{\lvert G \rvert}_{i=1}\alpha_i\hat{\boldsymbol{t}}_i$ with trilinear weights $\alpha_i \in [0,1]$ (cf. fig. \ref{fig:deformation_model}). The position $\boldsymbol{p}$ of the same vertex in the deformation field $\mathcal{V}$ can be described analogously by a weighted sum of deformed gridpoint positions $\ti$ as $\boldsymbol{p} = \sum^{\lvert G \rvert}_{i=1} \alpha_i \boldsymbol{t}_i$. We define an \textit{observation} $\mathcal{D}$ as an organized point cloud of the deformed object.

\subsection{Demonstration}
Most SotA approaches use either a low-resolution reference model of the deformable object \cite{Zollhofer.2014} or none at all \cite{Newcombe.2015, Innmann.2016}. To allow for user demonstrations of \acp{poi}, our algorithm requires a high-detail reference model to be created offline. To improve the stability of our solver, we limit the object's initial deformation with respect to its reference model by demonstrating a library $\mathcal{L}$ of reference models in an offline step. Each model in $\mathcal{L}$ is a triangle mesh of the object in a distinct deformation state. The models are generated by fusing several depth images into a \ac{tsdf} and then extracting the triangles via the MarchingCubes algorithm \cite{Lorensen.1987}. After the demonstration of the reference models, the user can select relevant \acp{poi} on the meshed surface via a graphical user interface. At runtime, after the first observation $\mathcal{D}$, the model in $\mathcal{L}$ most similar to $\mathcal{D}$ is used to initialize $G$.

\subsection{Correspondence identification}
The definition and identification of \textit{correspondences} links the current observation and the deformed reference model of the previous timestep $(t-1)$ and forms the basis of the deformation estimation: The estimation of a deformation is equivalent to the minimization of the distances between all correspondences. We define three correspondence types:
\paragraph{\Ac{p2p}}
Result from projecting each surface point $\pc$ of the deformed reference model into the image plane and comparing it to the corresponding point $\pca$ that has been measured. The quality of a \ac{p2p}-correspondence can be described by a weight $w_c = (\frac{w_d + w_n + w_v}{3})^2$ where $w_d$ denotes the distance between projected point $\boldsymbol{p}_c$ and its correspondent $\boldsymbol{p}^a_c$, $w_n$ the distance between $\boldsymbol{p}_c$'s normal $\boldsymbol{n}_c$ and its correspondent $\boldsymbol{n}^a_c$, and $\boldsymbol{w}_v$ the angle between the camera view direction $\boldsymbol{v}$ and $\boldsymbol{n}_c$.
\paragraph{\Ac{p2s}}
Projective correspondences such as \ac{p2p} typically only yield approximate, not exact, correspondences. \Ac{p2s}-correspondences add another degree of freedom by associating a point in the model to a \textit{plane} in the observation, defined by the \ac{p2p}-correspondence $\boldsymbol{p}^a_c$ and its normal $\boldsymbol{n}^a_c$. During deformation estimation, this allows to only minimize the distance $d_c$ along the normal.
\paragraph{Feature correspondences}
Unlike projective correspondences, correspondences based on feature matching can detect large deformations, tangential movements and rotations of the object out of the image plane. Prior work \cite{Guo.2016,Hansch.2014} and our own experiments have found the PFH and FPFH descriptors to be highly sensitive and specific but to scale poorly with the size of the point cloud, while SHOT descriptors scale linearly and are robust against outliers. We implement feature correspondences using SHOT, as its sensitivity suffices for most real-world applications.
\subsection{Deformation estimation}
The deformation of the reference object can be estimated by formulating an optimization problem to estimate their degrees of freedom and thus the deformation of the reference object. Using the notation introduced in \ref{sec:surface_and_deformation_model}, the deformation of a single grid point $i$ can be expressed as $\mathcal{V}_i = \hat{\boldsymbol{t}}_i - \boldsymbol{t}_i$. For estimating the deformation field, we split up all unknows into a single global rigid transformation $(\boldsymbol{t}, R)$ and many local transformations $(\ti, R_i)$ and combine them in a vector $\mathcal{X}$:
\begin{equation}
    \mathcal{X} = \left(
    \boldsymbol{t},~R, 
    \underbrace{\quad \ldots , \ti^{T} , \cdots \quad}_{3~\vert G \vert~translations}
    \vert
    \underbrace{\quad  \cdots , R_{i} , \cdots \quad}_{3~\vert G \vert~rotations}
    \right)^{T}
\end{equation}
The interpretation of correspondences as error terms $E$ allows to formulate the deformation estimation of $\mathcal{X}$ as an energy minimization problem, which is also suggested by \cite{Innmann.2016,Zollhofer.2014,Newcombe.2015}. This optimization can be regarded as a model regression problem and solved by existing solvers:
\begin{equation}
    \begin{aligned}
        E(\mathcal{X}) =~ &\omega_{p} E_{P\mathit{2}P}(\mathcal{X}) + \omega_{s} E_{P\mathit{2}S}(\mathcal{X})\\ 
            +~ &\omega_{f} E_{F}(\mathcal{X}) + \omega_{r} E_{Reg}(\mathcal{X})
    \end{aligned}
    \label{eqn:E_gesamt}
\end{equation}
\cite{Zollhofer.2014} and \cite{Innmann.2016} solve a similar high-dimensional nonlinear optimization problem by linearizing the model and using the Gauss-Newton method, incurring a significant overhead for the computation of the Jacobian $J$. \cite{Innmann.2016} splits the optimization into a two-stage process composed of a \textit{fixed registration} followed by a \textit{deformation estimation}. We leverage the fact observed in \cite{Sorkine.2007} that the deformation estimation can again be split into two independent sub-problems, which allows to solve for nonlinear rotations and linear translations using iterative Gauss-Newton on each subproblem in turn (``flip-flop'' strategy). We perform fixed registration, the estimation of a global transformation $(\boldsymbol{t}, R)$, via Prerejective RANSAC (PSC). For the deformation estimation, setting up the Jacobian for the error terms of the three correspondence types is straightforward:
\begin{equation}
    E_{P\mathit{2}P} = \sum^{\vert C \vert}_{c=1} \omega_{c} 
    ~ \Biggl| \Biggl|
        \underbrace{
            \underbrace{
                R \left[ \sum^{\vert G \vert}_{i=1} \alpha_{i}(\pcd)~\ti \right] + \boldsymbol{t}}_{\mathcal{V}(\pcd)} -\pca
        }_{\rVec_{P\mathit{2}P,C}(\mathcal{X})}
    \Biggl| \Biggl|^{2}_{2}
    \label{eqn:P2P_Error}
\end{equation}
\begin{figure}[H]
    \centering
    \begin{minipage}{.69\linewidth}
        \includegraphics[width=\linewidth]{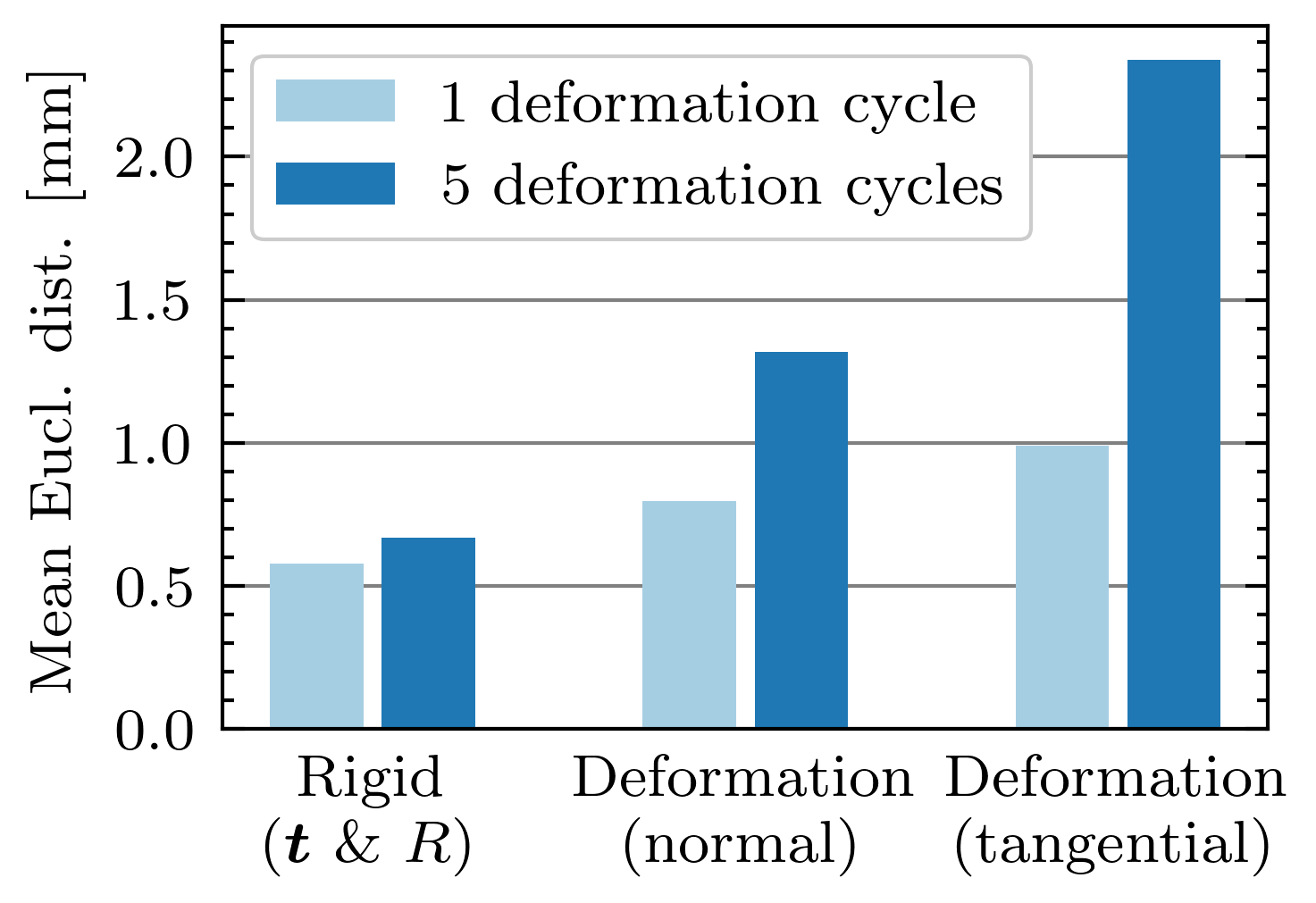}
    \end{minipage}%
    \hfill%
    \begin{minipage}{.31\linewidth}
        \begin{minipage}{\linewidth}
            \includegraphics[width=\linewidth]{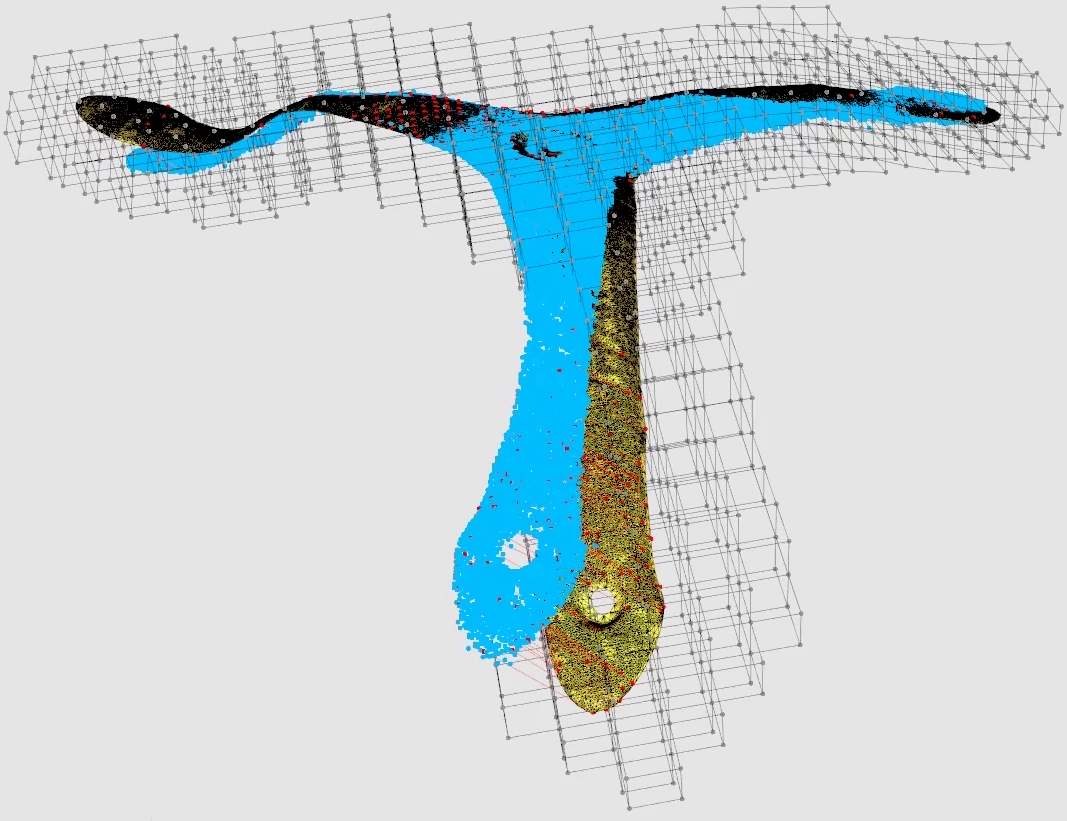}
        \end{minipage}\\
        \begin{minipage}{\linewidth}
            \includegraphics[width=\linewidth]{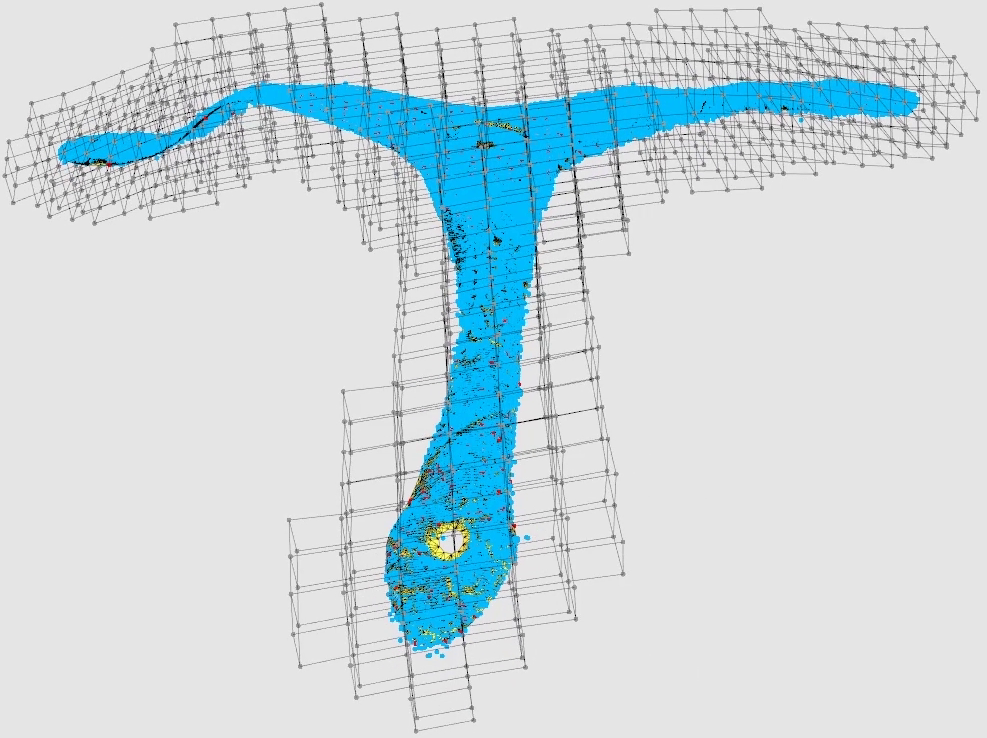}
        \end{minipage}
    \end{minipage}
    \caption{Precision for different deformation types (l.); estimated deformation grid and ground-truth point cloud (blue) before and after one deformation cycle (r.)}
    \label{fig:precision}
\end{figure}
\begin{equation}
    J_{P\mathit{2}P,ci} = \dfrac{\partial\left( 
        \omega_{p} ~\omega_{c} ~\rVec_{P\mathit{2}P,c}(\mathcal{X})\right)}
        {\partial \ti} = 
        \omega_{p} ~\omega_{c} ~\alpha_{i}(\pcd)
        \label{eqn:p2p_Jacobi}
\end{equation}
where $\lvert C \rvert$ is the number of correspondences and $J_{P\mathit{2}P,ci}$ is the entry at the $c^{th}$ row and $i^{th}$ column of the Jacobian $J_{P\mathit{2}P}$. The Jacobians for $E_{P\mathit{2}S}$ and $E_{F}$ can be found analogously.

\paragraph{Regularization}
With a single camera's perspective, it is impossible to observe the complete surface of an object. The spatial lack of correspondences implies an underdetermined equation system and $E$ being ill-conditioned. To alleviate this problem, we use an ARAP regularizer \cite{Sorkine.2007}, where non-observable surface points are deformed such that the total deformation of the body is \textit{as rigid as possible}. Unlike prior work \cite{Sorkine.2007,Sorkine.2017}, we estimate the deformation in terms of $\mathcal{V}$ instead of the mesh, leading to the adapted ARAP term
\begin{equation}
    E_{Reg} =
    \sum^{\vert G \vert}_{i=1} \sum_{j \in \mathcal{N}_{i}}
        \bigl| \bigl|
            \left(\ti - \tj \right) 
              - R_{i} \left(\tdi - \tdj \right) 
        \bigl| \bigl|^2_{2},
    \label{eqn:grid arap}
\end{equation}
where $\mathcal{N}_i$ denotes the \textit{neighborhood} (6 surrounding grid points) of grid point $i \in [1, \lvert G \rvert]$.
For each grid point $i$, our solver must solve for 6 unknowns describing its pose $(R_i | \ti)$. As shown in \cite{Sorkine.2017}, the (non-linear) estimation of $R_i$ can be solved in closed form given $\boldsymbol{t}_i$. For $\boldsymbol{t}_i$, we obtain
\begin{equation}
    \frac{\partial E_{Reg,i}}{\partial \ti} \overset{!}{=} 0 
 \Leftrightarrow
        \underbrace{	
            \sum_{j \in \mathcal{N}_{i}} (\ti - \tj)
        }_{L~\cdot~\mathcal{X}_{i}} = \sum_{j \in \mathcal{N}_{i}} \dfrac{R_{i}+R_{j}}{2} (\tdi - \tdj)
    \label{eqn:arap final}
    \end{equation}
where the left-hand side is the product of the Laplace matrix $L$ with the vector of all unknowns $\mathcal{X}_i$.
\paragraph{Flip-flop solver}
We iteratively estimate $R_i$ and $\boldsymbol{t}_i$ in turn by closed-form solving for $R_i$ via singular value decomposition (see \cite{Sorkine.2017} for details) and approximating $\boldsymbol{t}_i$ via Gauss-Newton, where the update step $\Delta \mathcal{X}$ is obtained via \ac{pcg}. Using the Jacobians derived above, we can obtain the deformation $\mathcal{X}_{t+1}$ after an update step via
\begin{align}
    J^{T}J &:= 
        J^{T}_{P\mathit{2}P}J_{P\mathit{2}P} + 
        J^{T}_{P\mathit{2}S}J_{P\mathit{2}S} + 
        J^{T}_{F}J_{F} + 
        L^{T}L \\
    J^{T}\boldsymbol{r} &:=
    J^{T}_{P\mathit{2}P} \boldsymbol{r}_{P\mathit{2}P} +
    J^{T}_{P\mathit{2}S} \boldsymbol{r}_{P\mathit{2}S} +
    J^{T}_{F} \boldsymbol{r}_{F} +
    L^{T}\boldsymbol{r}_{Reg}
    \label{eqn:pcg LHS}
\end{align}
\begin{align}
    J^{T}J \,\ \boldsymbol{\Delta} \mathcal{X} &= J^{T}\boldsymbol{r}\\
    \mathcal{X}_{t+1} &= \mathcal{X}_{t} + \boldsymbol{\Delta} \mathcal{X}
    \label{eqn:pcg}
\end{align}

\begin{figure}
    \centering
    \begin{minipage}{.65\linewidth}
        \includegraphics[width=\linewidth]{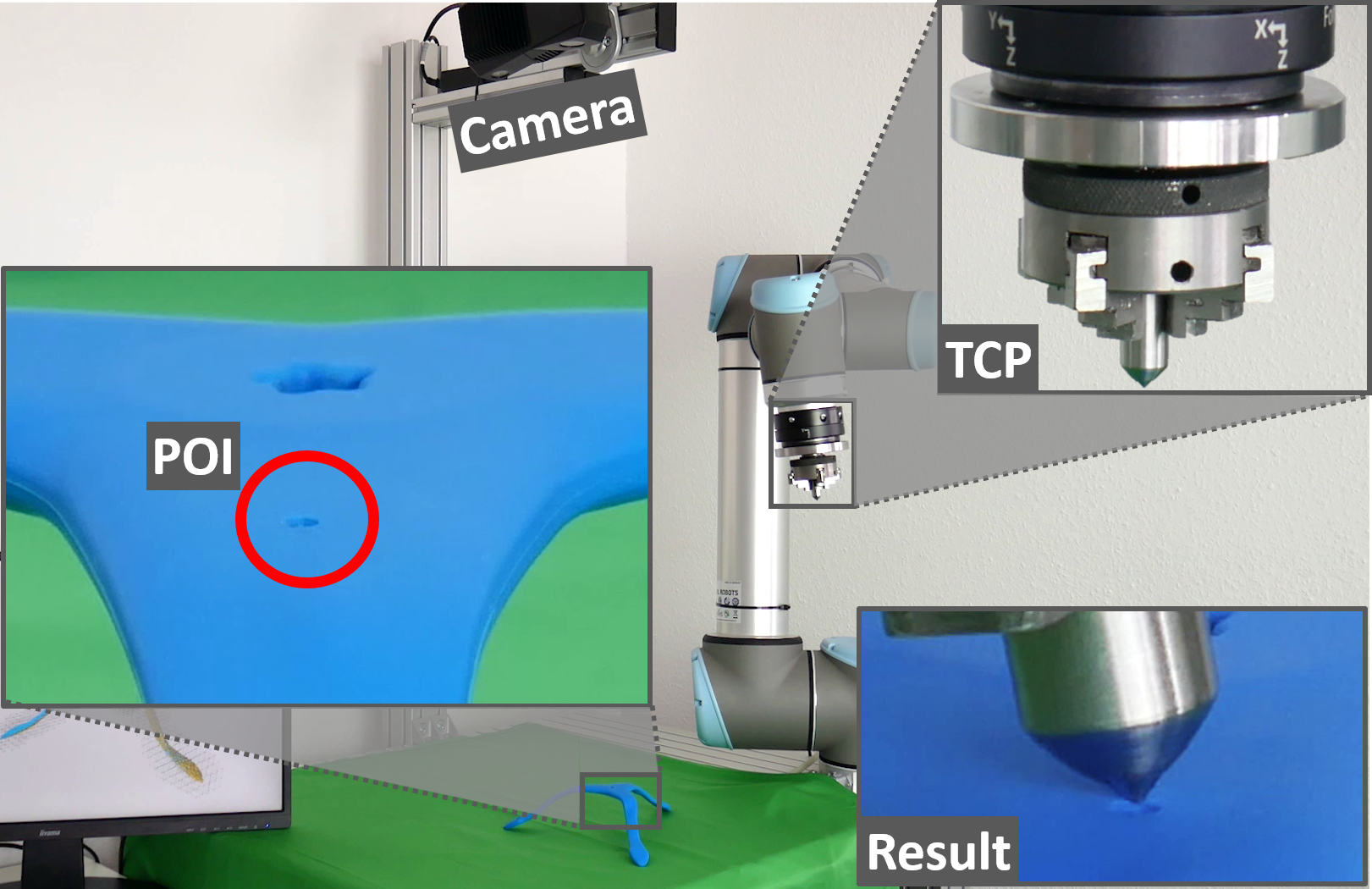}    
    \end{minipage}%
    \hfill
    \begin{minipage}{.335\linewidth}
    \includegraphics[width=\linewidth]{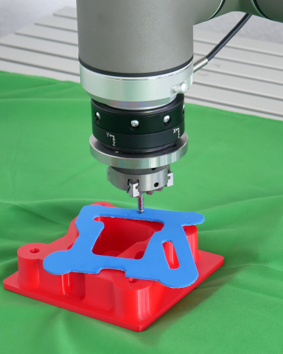}        
    \end{minipage}
    \caption{\ac{poi} tracking on a deformable tripod (l.); rubber seal assembly (r.)}
    \label{fig:experiments}
\end{figure}
\section{Results}
\paragraph{Precision} \label{par:precision}In a first set of experiments, we assess the precision of our approach by comparing tracking results versus manually labeled correspondences. 10 \acp{poi} on a deformable tripod were considered, with each \ac{poi} also fitted with a color-coded marker to facilitate manual labeling.\footnote{Since our algorithm only considers geometric features, the presence of the markers neither helped nor hurt the algorithm.} The tripod was repeatedly deformed and the poses of the \acp{poi} were estimated by our algorithm as well as via the markers (cf. fig. \ref{fig:precision}). Our approach was capable of \textit{localizing} all \acp{poi} with sub-millimeter accuracy, and \textit{tracking} all \acp{poi} with errors between 0.6 and 2.2 mm. Unlike feature-matching based approaches, we always estimate the deformation of the complete surface and thereby avoid ``mismatching'' \acp{poi} by design. 
\paragraph{Performance} A significant advantage of our approach is that each step of the solver pipeline can be efficiently parallelized. We benchmarked our algorithm using reference and deformation models at fine\footnote{Reference model: 30000 vertices, $\mathcal{V}$: 3250 grid points} and coarse\footnote{Reference model: 15000 vertices, $\mathcal{V}$: 700 grid points} resolutions. A parallelized CPU implementation of our algorithm localized all \acp{poi} in under $1.5s$ in both cases on consumer hardware.
\paragraph{Robot experiments} In a first robot experiment, we track a point on the surface of a tripod subjected to several deformations of up to 20\% of the tripod's arm length, or ca. 2.5 cm. We use a UR5 robot equipped with a measuring tip to visualize tracking results (cf. fig. \ref{fig:experiments} (l.)), confirming  precision within 2 mm. In a second experiment, we use our approach to position a flat rubber seal on a housing, illustrating its potential for real-world industrial applications (cf. fig. \ref{fig:experiments} (r.)).

\section{Discussion and Outlook}
Our approach and solver pipeline allows efficient tracking and localization of \acp{poi} on deformable objects. Where prior work requires markers, explicit modelling or does not allow for offline \ac{poi} definition, our approach achieves sub-millimeter precision localization and millimeter-precision tracking without these drawbacks. This makes it particularly suitable for applications in industrial robotics and flexible, quickly reconfigurable assembly or surface treatment tasks. We are working on integrating our solution into an industrial robot manipulation framework, a more efficient GPU implementation and a more extensive evaluation on a wider set of benchmarks.
\bibliographystyle{IEEEtran}
\bibliography{deformable_objects.bib}

\end{document}